\documentclass{article}
\usepackage{ijcai19}

\usepackage{times}
\usepackage{soul}
\usepackage{url}
\usepackage[hidelinks]{hyperref}
\usepackage[utf8]{inputenc}
\usepackage[small]{caption}
\usepackage{graphicx}
\usepackage{amsmath}
\usepackage{amssymb}
\usepackage{booktabs}
\usepackage{algorithm}
\usepackage{algpseudocode}
\usepackage{booktabs}
\graphicspath{{figures/}}

\usepackage{mathtools}
\DeclarePairedDelimiter{\norm}{\lVert}{\rVert}

\usepackage{amsthm}
\newtheorem{mydefinition}{Definition}

\newtheorem{theorem}[mydefinition]{Theorem}

\newcommand{\mbs}[1]{\ensuremath{\boldsymbol{#1}}}

\newcommand{\cX}{\mathcal{X}}

\newcommand{\ex}{\mathbb{E}}

\newcommand{\RR}{\mathbb{R}}

\newcommand{\thetav}{\mbs{\theta}}

\newcommand{\kv}{\mathbf{k}}

\newcommand{\x}{\mathbf{x}}
\newcommand{\y}{\mathbf{y}}

\newcommand{\K}{\mathbf{K}}

\newcommand{\X}{\mathbf{X}}

\newcommand{\xt}{\mathbf{x}_t}

\newcommand{\secref}[1]{\hyperref[#1]{Section~\ref{#1}}}
\newcommand{\figref}[1]{\hyperref[#1]{Figure~\ref{#1}}}
\newcommand{\algoref}[1]{\hyperref[#1]{Algorithm~\ref{#1}}}

\title{Fully Distributed Bayesian Optimization with Stochastic Policies}

\author{
Javier Garcia-Barcos$^1$
\And
Ruben Martinez-Cantin$^{1,2}$
\affiliations
$^1$Instituto de Investigacion en Ingenieria de Aragon, University of Zaragoza\\
$^2$Centro Universitario de la Defensa, Zaragoza
\emails
{jgbarcos, rmcantin}@unizar.es
}

\begin{document}

\maketitle

\begin{abstract}
Bayesian optimization has become a popular method for high-throughput computing, like the design of computer experiments or hyperparameter tuning of expensive models, where sample efficiency is mandatory. In these applications, distributed and scalable architectures are a necessity. However, Bayesian optimization is mostly sequential. Even parallel variants require certain computations between samples, limiting the parallelization bandwidth. Thompson sampling has been previously applied for distributed Bayesian optimization. But, when compared with other acquisition functions in the sequential setting, Thompson sampling is known to perform suboptimally. In this paper, we present a new method for fully distributed Bayesian optimization, which can be combined with any acquisition function. Our approach considers Bayesian optimization as a partially observable Markov decision process. In this context, stochastic policies, such as the Boltzmann policy, have some interesting properties which can also be studied for Bayesian optimization. Furthermore, the Boltzmann policy trivially allows a distributed Bayesian optimization implementation with high level of parallelism and scalability. We present results in several benchmarks and applications that show the performance of our method.

\end{abstract}

\section{Introduction}
Many engineering problems and scientific phenomena are being studied in high-throughput computing facilities, through complex computer models or simulators. Due to the large amount of resources that are needed, experiments should be carefully selected and studied. This is known as the design and analysis of computer experiments \cite{Sacks89SS}. Bayesian optimization (BO) can be used for designing computer experiments which require the search of an optimum value \cite{Jones:1998,Mockus78}. BO carefully selects the next experiments to perform in order to find the optimum value as efficiently as possible. For example, consider the problem of finding the optimal shape of a wing profile to reduce the drag force. Instead of dealing with the infamous equations of Navier-Stokes, we can experiment with a Computational Fluid Dynamics (CFD) simulator and check the outcome \cite{forrester2006optimization,martinez2018funneled}. Thus, for the optimization algorithm, the simulator becomes a black-box where we only care about the resulting drag force. For scientific experiments,  we may want to adjust the parameters of a computational model of cell migration to mimic the behaviour of \emph{in vivo} or \emph{in vitro} experiments \cite{merino2018integration}, or we may want to find new drugs through virtual screening \cite{hernandez2017parallel}. Finally, one of the most popular applications for BO is the tuning of hyperparameters of complex machine learning models, such as deep neural networks \cite{Snoek2012,klein-aistats17}.

BO achieves sample efficiency by learning a probabilistic surrogate model of the the target function. Then, using several heuristics, called acquisition functions, we can select the next point to be evaluated or experimented. Therefore, BO is intrinsically a sequential process. Each new observation is incorporated into the surrogate model, which at the same time, modifies the acquisition function for the selection of future experiments. There are certain algorithms that allow parallel or batch queries of experiments \cite{Snoek2012,gonzalez2016batch,desautels2014parallelizing}, but they still require the queries to be computed in a sequential manner. In high-throughput systems, these methods require a central node that computes and dispatches the queries. Furthermore, as pointed out by \cite{kandasamy:par-ts}, many of those methods do not allow for asynchronous execution. To the authors knowledge, only the Thompson sampling approach from \cite{hernandez2017parallel} can be fully distributed. However, it is well-known that there are many other acquisition functions that performs better than Thompson sampling in practice \cite{shahriari2016taking}. In this work, we introduce a new method to allow a fully distributed BO, which can be combined with any acquisition function.

However, the contribution of our paper is twofold. Before introducing our distributed BO algorithm, we present a new portrayal of BO in the Markov decision process framework, following the analysis of \cite{ToussaintBSG}. We show how this new framework allows a greater understanding of the features and capabilities of BO. Then, we introduce the idea of stochastic policies for BO, which is the ingredient necessary to perform fully distributed BO in combination with any acquisition function. Furthermore, we analyze the advantages of the stochastic policies beyond parallelization, from a theoretical and practical point of view.

\section{The Optimization Agent}
We consider the optimization algorithm as an agent that is interacting with the environment, which is the space of functions. The current state of the environment is the target function $f$. The agent does not have full observability of the state and can only perform partial observations by querying the function $y_t = f(\xt)$. As pointed out in \cite{ToussaintBSG}, the future decisions made by the agent can be modelled as a \emph{partially observable Markov decision process} (POMDP). A similar analogy, although from a control theory perspective is also presented in \cite{lam2016bayesian}.

\subsection{POMDPs\label{sec:pomdp}}
A (PO)MDP is a stochastic model where the agent and the environment are fully represented by state variables $s_t \in \mathcal{S}$ and the agent can change that state by performing actions $a_t \in \mathcal{A}$ in a Markovian way $p(s_{t+1} | a_t, s_t)$.  The (PO)MDP model also assumes that the agent is rational, acting to maximize the future expected reward $\ex[\sum_{t=0}^N R(s_t, a_t)]$. The behavior of the agent is encoded in the policy which maps states to actions $a_t = \pi(s_t)$. In order to rank the possible actions at one state, we can compute the Q-function which represents the \emph{quality} of taking a certain action considering the future reward. The optimal Q-function $Q^*$ can be computed by doing \emph{full backups} of future rewards and actions recursively. Then, we can obtain the optimal greedy function by:
\begin{equation}
  \label{eq:greedypolicy}
 \pi^*(s_t) = \arg \max_a Q^*(s_t, a_t) 
\end{equation}

In the POMDP setting, the agent does not have full observability of the state and must rely on partial observations $y_t$ following an observation model $p(y_t | s_t, a_t)$. Thus, the agent relies on \emph{beliefs}, which are the distributions over possible states, given the known observations and actions $b_t = p(s_t | a_{0:t}, y_{0:t})$. Because, the belief is a sufficient statistic, it can be shown that a POMDP on state space, is equivalent to a MDP on belief space \cite{Kaelbling96jmlr}. In this case, the transition model becomes $p(b_{t+1} | a_t, b_t) = \int_y p(b_{t+1} | a_t, b_t, y_t) p(y_t|a_t,b_t) dy$. Then, the reward and the policy become functions of the belief $r(b_t,a_t) = \int_s b_t(s) R(s_t,a_t) ds$ and $\pi(b_t) = \int_s b_t(s) \pi(s_t) ds$. In this case, the optimal policy and Q-function can also be mapped to the belief space $\pi^*(b_t) = \arg \max_a Q^*(b_t, a_t)$.

\begin{table}
\centering
\begin{tabular}{lr}  
\toprule
POMDP / belief MDP & Bayesian optimization \\
\midrule
State: $s_t$ & Target function: $f$ \\
Action: $a_t$ & Next query: $x_{t+1}$       \\
Observation: $y_t$ & Response value: $y_t = f(x_t)$  \\
Belief: $b_t=p(s_t)$ & Surrogate model: $p(f)$      \\
Q-function: $Q^*(b_t,a_t)$ & Acquisition function: $\alpha(x,p(f))$      \\
Reward: $R(s_t,a_t)$ & Improvement: $\max(0,y_{t+1} - \rho_t)$\\
\bottomrule
\end{tabular}
\caption{Comparison of POMDP and belief MPD terms with respect to the corresponding elements in BO}
\label{tab:comparison}
\end{table}

\subsection{Bayesian Optimization\label{sec:bayesianoptimization}}
Bayesian optimization is a set of optimization methods~\cite{shahriari2016taking} with two important distinct features: a \emph{probabilistic surrogate model} $p(f)$ to learn the properties and features of the target function that we are trying to optimize and, an \emph{acquisition function} $\alpha(x,p(f))$ that, based on the surrogate model, rate the potential interest of subsequent queries.

More formally, BO tries to optimize a function $f:\cX\to\RR$ over some domain $\cX\subset\RR^d$, by carefully selecting the queries of the function to reduce the number of evaluations of $f$ before finding the optimum $x^*$. At iteration $t$, all previously observed values $\y=y_{1:t}$ at queried points $\X=\x_{1:t}$ are used to construct a probabilistic surrogate model $p(f|y_{1:t},\x_{1:t})$. Typically, the next query location $\x_{t+1}$ is determined by greedily optimizing the acquisition function in $\cX$:
\begin{equation}
  \label{eq:acquisition}
 \x_{t+1} = \arg \max_{\x \in \cX} \alpha\left(\x, p(f\;|\;y_{1:t},\x_{1:t})\right)
\end{equation}
For example, we can use the expected improvement (EI) as the acquisition function \cite{Mockus78}:
\begin{equation}
	\label{eq:ei}
	EI_t(\x) = \mathbb{E}_{p(y_{t+1}\;|\;y_{1:t},\x_{1:t})} \left[\max(0,y_{t+1} - \rho_t)\right],
\end{equation}
where $\rho_t=\max(y_1,\ldots,y_t)$ is the incumbent optimum at that iteration. EI is still one of the most popular choices, although there are multiple alternatives depending on the criteria selected, such as \emph{optimism in the face of uncertainty} \cite{Srinivas10}, information about the optimum \cite{HennigSchuler2012,NIPS2014_5324,pmlr-v70-wang17e}, etc. 

It has been found that EI might be unstable in the first iterations due to the lack of information \cite{Jones:1998,Bull2011}. Therefore, the
optimization is initialized with $p$ evaluations by sampling from low discrepancy sequences.

\subsubsection{Surrogate Model\label{sec:gp}}
Most frequently, this takes the form of a Gaussian process (GP), although other alternatives have been presented, such as Bayesian neural networks~\cite{hernandez2017parallel}. For the remainder of the paper we consider a GP with zero mean and kernel $k:\cX\times\cX\to\RR$ as the surrogate model. The kernel is chosen to be the Mat\'ern kernel with smoothness parameter $\nu$ and hyperparameters $\thetav$. The GP posterior model gives predictions at a query point $\x_q$ which are normally distributed $y_q \sim \mathcal{N}(\mu(\x_q), \sigma^2(\x_q))$, such that $\mu(\x_q) = \kv(\x_q)^T\K^{-1}\y$, and $ \sigma ^2 (\x_q) = k(\x_q, \x_q) - \kv(\x_q)^T \K^{-1} \kv(\x_q)$ where 
$\kv(\x_q) = \left[k(\x_q,\x_i)\right]_{\x_i \in \X}$ and $\K = \left[\kv(\x_i,\x_j)\right]_{\x_i,\x_j \in \X} + \mathbf{I}\sigma^2_n$. 


\subsection{BO as a POMDP}
As can be seen, Equation \eqref{eq:greedypolicy} is analogous to Equation \eqref{eq:acquisition}. The POMDP framework, as a rational Bayesian model, can also be applied to BO. Table \ref{tab:comparison} summarizes the connections between those frameworks. If we use EI as the acquisition function, then we assume that the improvement function $I=\max(0,y_{t+1} - \rho_t)$ is the reward. In this case, the EI is the optimal myopic policy for POMDP, as it maximizes the expected reward one step ahead. More interestingly, entropy based acquisition functions can be interpreted as an active learning problem, for which the POMDP can also be applied as a framework \cite{lopes2014active}.

\section{Stochastic Policies for BO\label{sec:stochasticpolicy}}

BO policies are typically greedy in two ways. First, they are temporally greedy, that is, they look for the maximum immediate reward, although some lookahead alternatives have been proposed in the past \cite{gonzalez2016glasses,lam2016bayesian}. Second, they are spatially greedy, that is, they only select the action or next query that maximizes the acquisition function (or Q-function). However, one might want to explore suboptimal actions to gather knowledge about the world. In fact, Q-learning algorithms may not converge to the solution using a greedy policy due to the lack of infinite exploration.

\begin{figure}
\centering
    \includegraphics[width=0.9\linewidth]{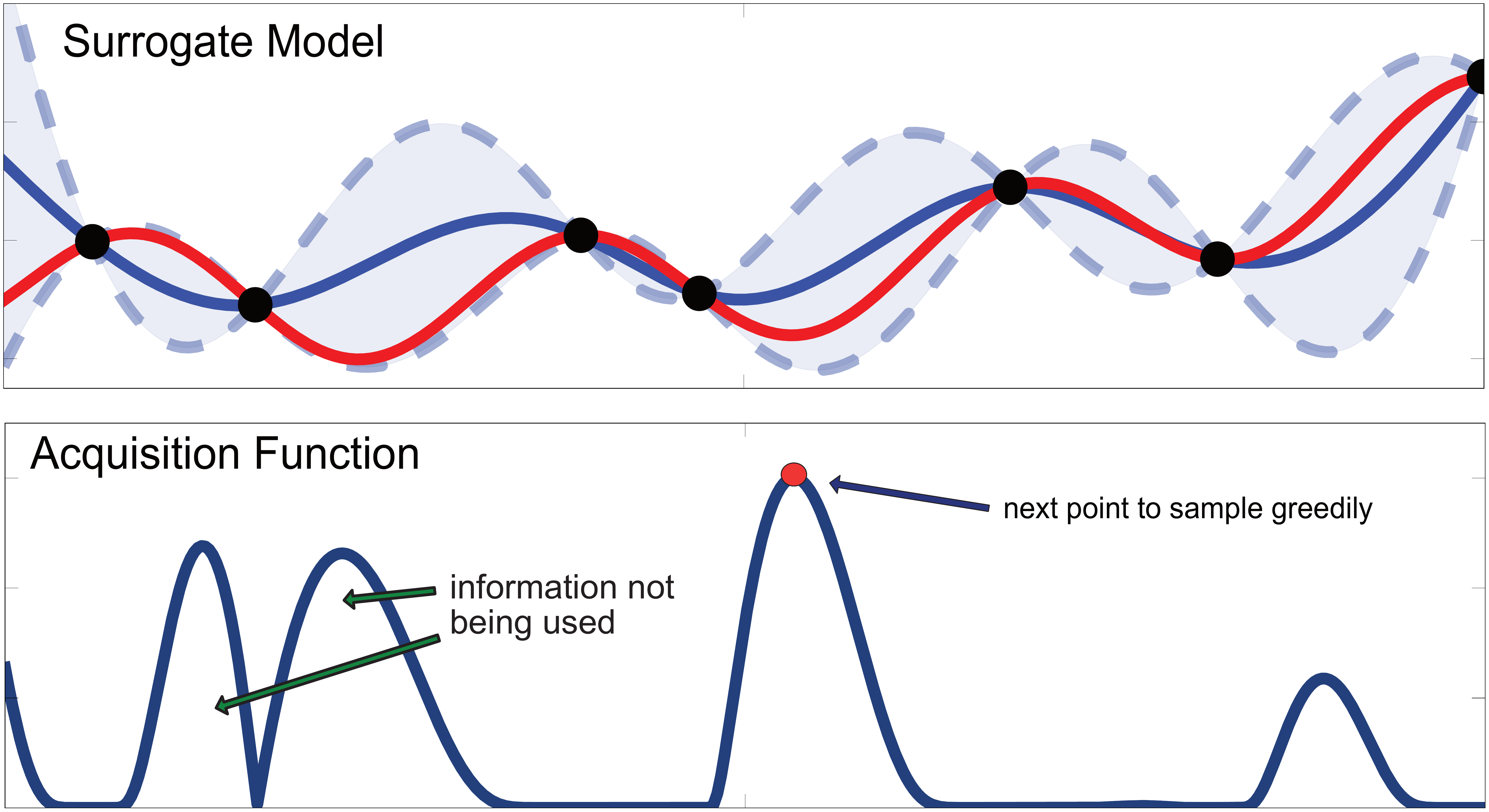}
    \caption{Example of a surrogate model (blue) on top of the target function (red). The greedy policy selects the maximum of the acquisition function (red dot), but completely ignores the rest of the regions which are almost as valuable.}
    \label{fig:example}
\end{figure}

Many acquisition functions, such as those mentioned in Section \ref{sec:bayesianoptimization} are already designed to trade-off exploration and exploitation. However, this trade-off is based on the assumption that the surrogate model \emph{is good enough} to predict both the expected function value and its uncertainty. In practice, even if the model is chosen carefully, we still need to learn the hyperparameters. Thus, for the first few iterations, the model is inaccurate. Interestingly enough, theoretical results rely on bounding the length-scales of the GP kernels to artificially increase the exploration, or directly using $\epsilon$-greedy strategies to guarantee near-optimal convergence rates with unknown hyperparameters \cite{wang2014theoretical,Bull2011}. For inaccurate models, the acquisition function is still able to provide some information, but by greedily selecting a single value we are wasting some of that information. Figure \ref{fig:example} shows how the greedy policy only cares about the central mode of the acquisition function, while the two modes on the left are almost as interesting to be explored.

Instead, we propose to use a stochastic policy such as the following Boltzmann policy (also known as Gibbs or softmax policy):
\begin{equation}
  \label{eq:sp}
  p(\x_{t+1}\;|\;y_{1:t}, \x_{1:t}) = \frac{e^{\beta_t \alpha\left(x_{t+1}, p(f\;|\;y_{1:t},\x_{1:t})\right)}} {\int_{x\in\cX} e^{\beta_t \alpha\left(x, p(f\;|\;y_{1:t},\x_{1:t})\right)} dx}
\end{equation}
This policy defines a probability distribution for the next query or action. Thus, the actual next query is selected by sampling that distribution $\x_{t+1} \sim   p(\x_{t+1}\;|\;y_{1:t}, \x_{1:t})$. This policy allows exploration even if the model is completely biased. Furthermore, it has some interesting properties for BO convergence as we will discuss in Section \ref{sec:theory}. The main result of this work is that the sampling process of $\x_{t+1}$ can be done \textbf{in parallel and fully distributed}, as will be discussed in Section \ref{sec:distributed}. This approach can be applied to any acquisition function or surrogate model that can be found in the literature. Nevertheless, the theoretical analysis and posterior experimentation focuses on GP and EI as previously discussed in Section \ref{sec:bayesianoptimization}.

\subsection{Theoretical Analysis\label{sec:theory}}
Convergence of BO algorithms has been extensively studied in terms of convergence rates \cite{Bull2011} or regret bounds in a bandit setting \cite{Srinivas10,wang2014theoretical}. In this paper we are going to follow upon the analysis by \cite{Bull2011} for the expected improvement. We show how the stochastic policy from equation \ref{eq:sp} has the same rates as the greedy policy in the limiting case. We also show that the stochastic policy does not need to rely on the $\epsilon$-greedy strategy for near-optimal rates. 

Let $\cX \subset \RR^d$ be compact with non-empty interior. For a function $f :\cX \to \RR$, let $\ex_f^u$ denote the expectation when minimizing the fixed function $f$ with strategy $u$, as $u$ can be random.  We assume a prior $\pi$ for $p(f)$, following a Gaussian process with a Mat\'ern kernel with smoothness parameter $\nu$ and length-scales $\thetav$, having the Square exponential kernel as the limiting case of $\nu \rightarrow \infty$. Each kernel $K_\theta$ is associated with a space of functions $\mathcal{H}_\theta(X)$, its reproducing-kernel Hilbert space (RKHS). 

\begin{mydefinition}
  \label{def:ei-sp}
  An $EI(\pi,\beta_t)$ strategy chooses:
    \begin{enumerate}
    \item initial design points $\x_1, \dots, \x_k$ independently of $f$; and
    \item further design points $\x_{t+1}\ (t \ge k)$ sampled from \eqref{eq:sp}.
  \end{enumerate}
\end{mydefinition}
This is analogous to \cite[Def. 1]{Bull2011}, but replacing the greedy selection by the stochastic selection. The $EI(\pi,\beta_t)$ strategy can also be adapted to consider estimated parameters \cite[Def. 3]{Bull2011}. Note that, for the stochastic policy, there is no need to enforce that the selection is dense for constant values of the acquisition function.

\begin{theorem}
  \label{def:glie}
  Let $\cX$ be a finite space. If $\beta_t = \ln t / C_t$ and $C_t = \max_x | \max_z \alpha(z,p(f)) - \alpha(x,p(f))|$, then \eqref{eq:sp} is a greedy in the limit with infinite exploration (GLIE) policy. Therefore:
  \begin{enumerate}
  \item each point $\x$ is queried infinitely often if we use the $EI(\pi,\beta_t)$ strategy infinitely often, and
  \item in the limit, the $EI(\pi,\beta_t)$ policy is greedy with respect to the acquisition function $\alpha(\cdot)$ with probability 1.
  \end{enumerate}
\end{theorem}
The proof can be found in \cite[Ap. B]{singh2000convergence} following the relations from Table \ref{tab:comparison}. In order to generalize the previous result to $\cX \in \RR^d$, we can partition $\cX$ in $n$ regions of size $\mathcal{O}(t^{1/d})$. Following $EI(\pi,\beta_t)$ and assuming a large $t$ each region will be sampled with high probability, thus, the mesh norm is small, which is the requirement for near-optimal rates \cite[Lemma 12]{Bull2011}. In fact, with $\beta_1$ the sampling distribution is exactly uniform.

\begin{theorem}
  \label{thm:ei-sp-optimal}

  Let $EI(\pi,\beta_t)$ be the strategy in Definition \ref{def:ei-sp}.  If $\nu < \infty$, then for any $R > 0$,
  \[\sup_{\norm{f}_{\mathcal{H}_\theta(X)} \le R} \ex_f^u[f(x_t^*)-\min f] = O( (t/\log t)^{-\nu/d}(\log t)^{\alpha}),\]
  with probability 1, while if $\nu = \infty$, the statement holds for all $\nu < \infty$.
\end{theorem}
The sketch of the proof is based on \cite[Theorem 5]{Bull2011}, which uses greedy EI for optimality combined with $\epsilon$-greedy to guarantee the reduced mesh norm. Following Theorem \ref{def:glie}, for large $t$, the mesh norm will be small and the strategy will be greedy with probability 1. We encourage the reader to follow up the discussion in \cite{singh2000convergence} where it is shown that, for certain values of $\epsilon$, the strategy followed by \cite{Bull2011} is also GLIE.

\begin{algorithm}[b]
\renewcommand{\algorithmicrequire}{\textbf{Input:}}
\caption{BO-NODE}\label{al:bonode}
\begin{algorithmic}[1]
\Require Budget $T$, low discrepancy sequence $LD$.
\State Query $LD$ for $p$ initialization points based on node id.
\State Broadcast $\x_{1:p}, y_{1:p}$
\For{$t = p \ldots T$}  
   \State Collect $\x, y$ from other nodes when available.
   \State Update surrogate model $p(f|\x_{1:t},y_{1:t})$  
   \State Sample $\x_{t+1}$ with Equation \eqref{eq:sp}.
   \State Broadcast $\x_{t+1}$ and $y_{t+1} = f(x_{t+1})$
\EndFor
\end{algorithmic}
\end{algorithm}

\subsection{Distributed BO\label{sec:distributed}}
Now we are ready to present the main contribution of this work. The stochastic policy presented in Section \ref{sec:stochasticpolicy} allow us to implement BO in a fully distributed setup which can be easily deployed and scaled. Contrary to most parallel or batch BO methods which require of a central node to keep track of the queries computed and deployed, our approach does not need a centralized node. All the computation can be done in each node of the distributed system. Furthermore, for optimal results, the nodes only need to broadcast their queries and observed values $\{\x_{t}, y_t\}$, requiring minimal communication bandwidth. In addition to that, communication can be asynchronous and be even robust to failures in the network, as the order of the queries and observations is irrelevant.

Algorithm \ref{al:bonode} summarizes the code to be deployed in each node of the computing cluster or distributed system. The initialization part requires sampling from a low discrepancy sequence. This can be easily distributed by setting the precomputed sequence on a lookup table where each node accesses it based on their id. Once the initialization phase is done, each node builds its own surrogate model (e.g.: a GP) with all the data that is available to them. There is no requirement for the models to be synchronized or updated, although each node behaviour will be optimal if it has access to all the observations available as soon as possible.

\begin{figure}
    \centering
    \includegraphics[width=0.3\textwidth]{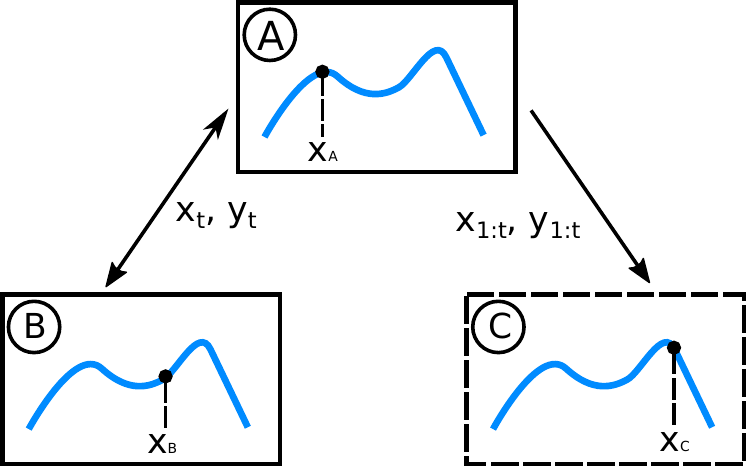}
    \caption{Visualization of the interaction between BO nodes. We have 3 nodes: A and B, which are already up and working; and C, a new node that we want to spin up mid-optimization. A and B only need to broadcast their new queries and observations. C needs to be given all the previous queries and observations up to the current instant $t$. Then it can resume its work in the same way as A and B.}
    \label{fig:network}
\end{figure}

One advantage of this setup is that it can be easily scaled by deploying new nodes using Algorithm \ref{al:bonode}, even halfway through the optimization process, as seen in Figure \ref{fig:network}. In that case, the first operation is to collect all the data that has been broadcasted in the network instead of using the low discrepancy sequence. Another advantage is that we can play with the $\beta_t$ parameter from equation \eqref{eq:sp} and deploy nodes more exploitative $\beta_t \rightarrow \infty$ or more exploratory $\beta_t \rightarrow 0$, or we can combine different kinds of node configurations, resulting in an overall behaviour analogous to \emph{parallel tempering} \cite{neal1996sampling}.

Many parallel BO methods have been proposed in the past few years. The main idea of all those methods is to ensure that the parallel experiments are well-distributed among the search space, which is problematic when we use a greedy approach without adding new information between queries. Thus, some authors include artificially augmented data by hallucinated observations \cite{ginsbourger2010kriging,Snoek2012} or by combining optimization with some degree of active learning in order maximize the knowledge about the target function \cite{desautels2014parallelizing,contal2013parallel,shah2015parallel} or by enforcing diversity through heuristics \cite{gonzalez2016batch}.



As noted on \cite{kandasamy:par-ts}, the majority of parallel methods are synchronous. Recently, a Thompson Sampling approach \cite{hernandez2017parallel,kandasamy:par-ts} has been applied to achieve fully distributed BO. To the authors knowledge, this is the only method comparable to our proposed distributed BO method.

\begin{figure*}
    \includegraphics[width=0.24\textwidth]{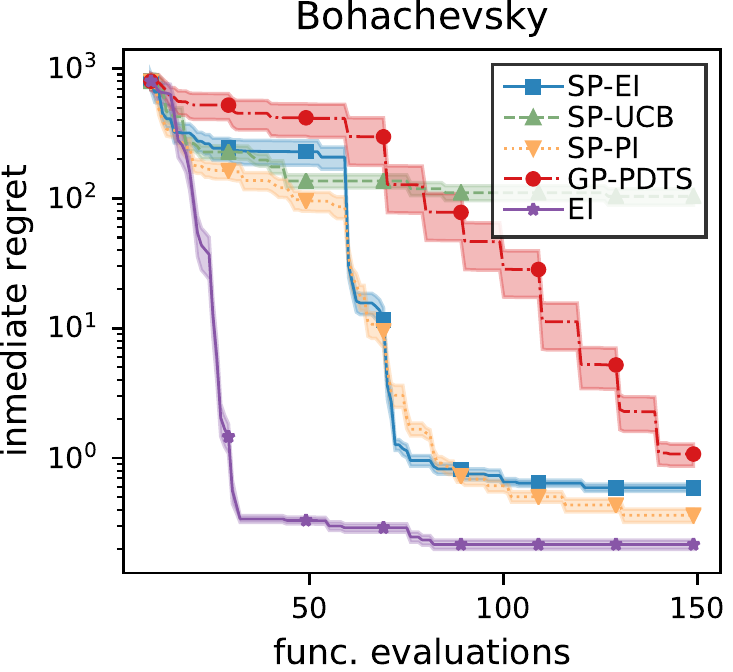}\hfill
    \includegraphics[width=0.24\textwidth]{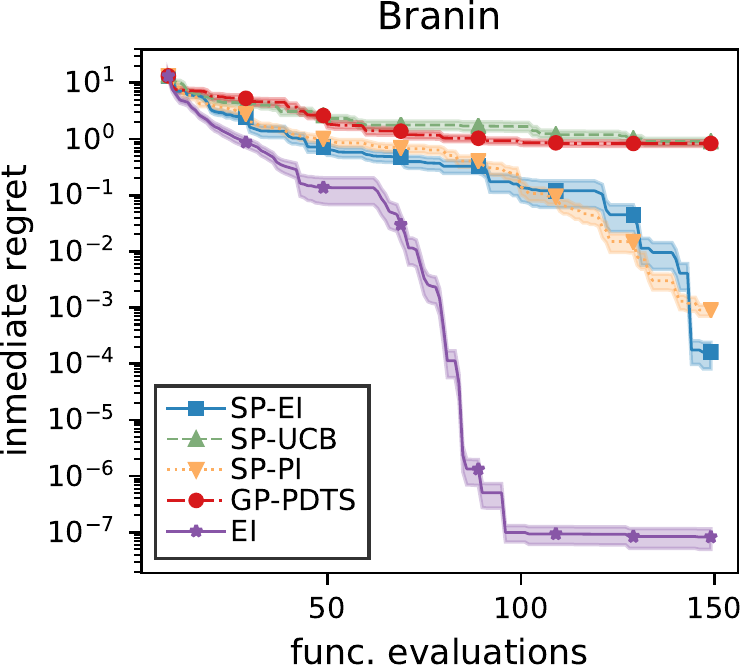}\hfill
    \includegraphics[width=0.24\textwidth]{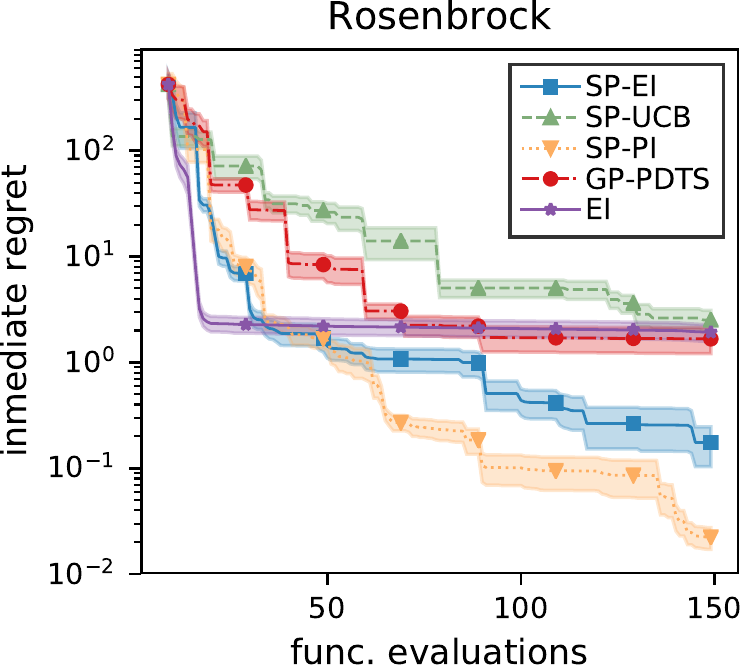}\hfill
    \includegraphics[width=0.24\textwidth]{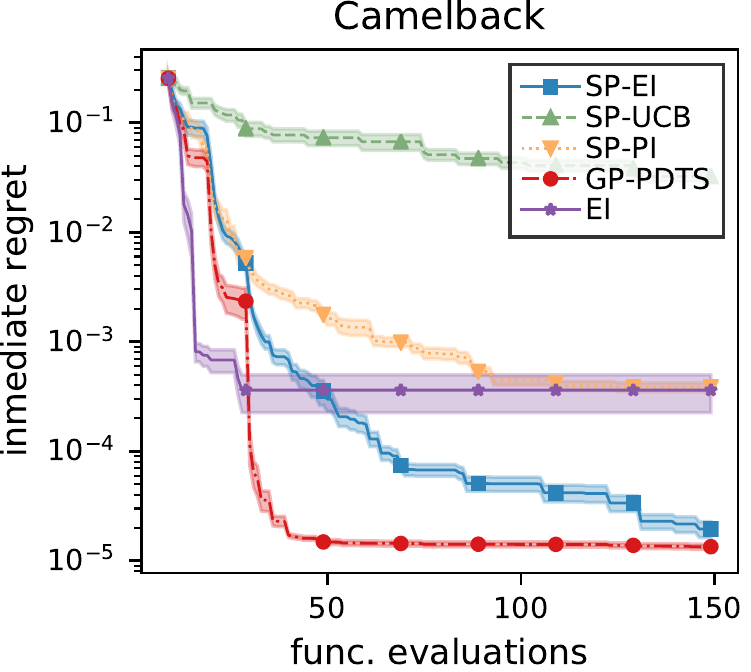}
    \includegraphics[width=0.24\textwidth]{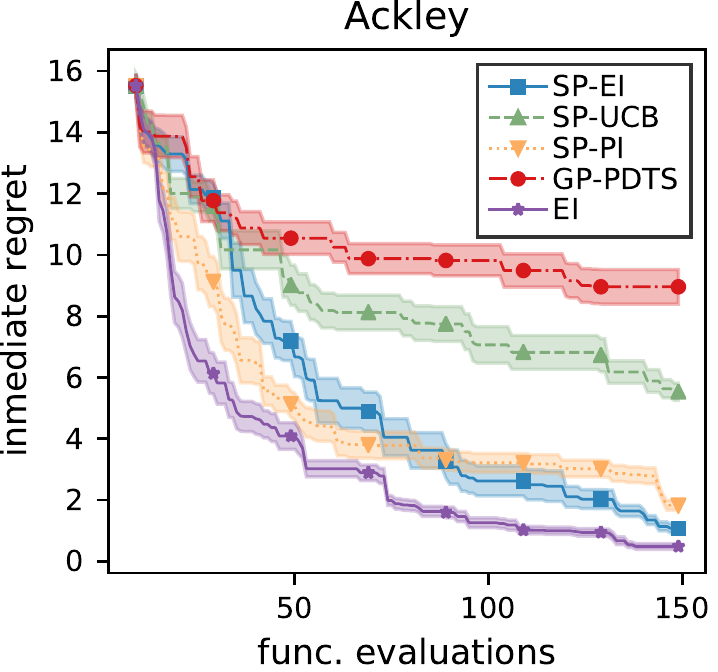}\hfill
    \includegraphics[width=0.24\textwidth]{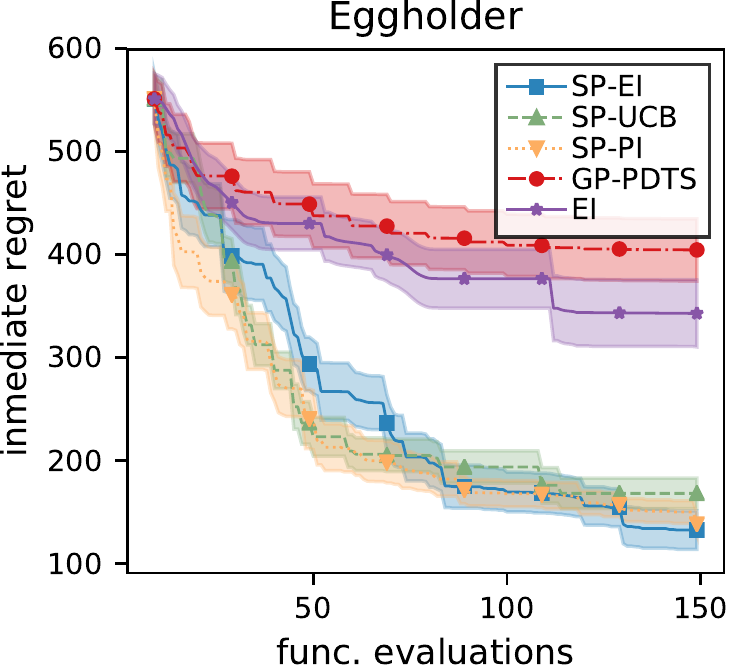}\hfill
    \includegraphics[width=0.24\textwidth]{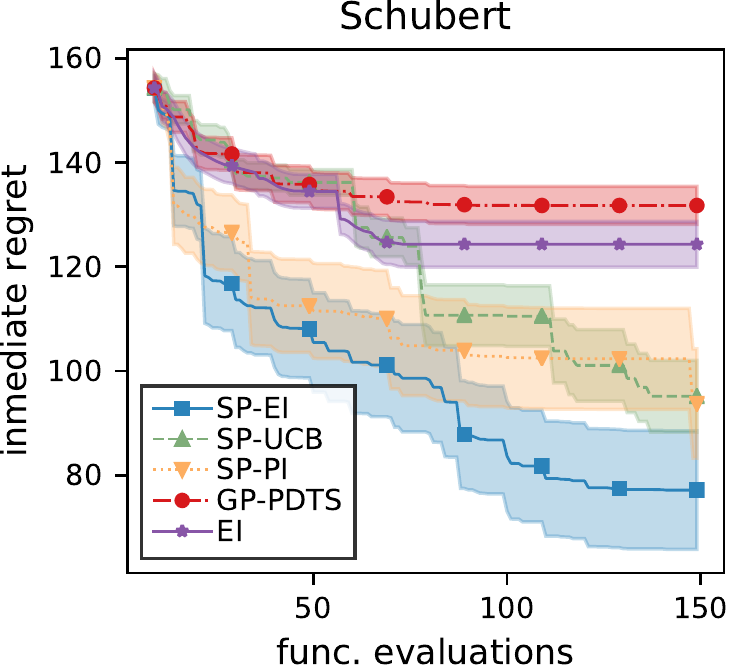}\hfill
    \includegraphics[width=0.24\textwidth]{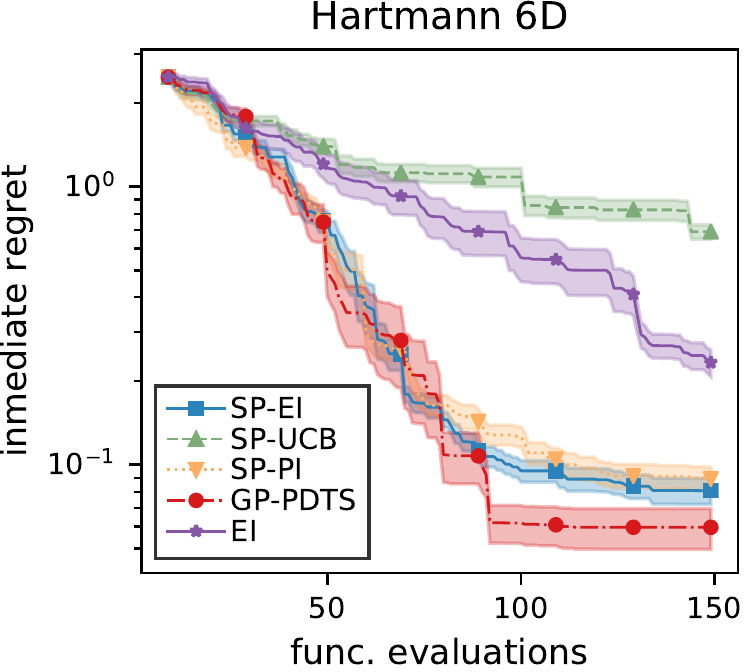}
    \caption{Inmediate regret on benchmark functions with the stochastic policy in combination with EI, UCB and PI. We include PDTS for comparison and sequential EI as baseline.}
    \label{benchmark}
\end{figure*}

\subsection{Sampling Strategies}
Sampling from equation \eqref{eq:sp} is not trivial. Most acquisition functions are highly multimodal with many large areas of low probability between modes. This is known to be problematic for MCMC methods. Tempering methods, such as simulated or parallel tempering \cite{neal1996sampling} usually perform better in these situations by allowing higher temperatures and, therefore, higher mixing of particles, in combination with lower temperatures to better represent the distribution. One advantage of our distribution is that it is already in Boltzmann form, allowing us to change the temperature by altering the $\beta_t$ parameter. In our case, since we have a target temperature that we want to use for sequential convergence (as seen in Theorem \ref{def:glie}) or for distributed exploration-exploitation (as seen in Section \ref{sec:distributed}), we can select the temperature profile in such a way that the lower temperature matches the target. Alternatively, we can select another profile and use importance sampling on the target temperature. In practice, we found that, for small dimensional problems as those typically found for BO, a mixture of Gaussians with different variance as the proposal function and a fixed temperature using Metropolis-Hasting can be enough to get a good distribution.

\section{Discussion}
Although the main result of this work is on distributed BO, the POMDP framework can introduce new interesting ideas to explore in the context of BO. For example, multitasking or multifidelity systems have a great potential for high-throughput computing with BO, where we are able to improve the convergence of our method by introducing new information from other sources, either from related problems where we already have previous experiments or from less expensive sources like a smaller training datasets, incomplete executions or a simpler simulator. In both scenarios the target function $f$ now belongs to a family of functions from other tasks or fidelities $f_k$. In the bandit setup, this is equivalent to contextual bandits. In the POMDP formulation, the function $f$ is analogous to the state $s$ (see Table \ref{tab:comparison}). The POMDP model includes a transition function $p(s_{t+1}\;|\;a_t,s_t)$, which can be known a priori or learned, that defines a probability distribution of transitions between states. This transition function allows more flexibility than contextual variables.

Recent results on lookahead policies try to avoid the temporal greediness of BO. One interesting approach is based on dynamic programming \cite{lam2016bayesian}. However, as pointed out by the authors, the dynamic programming approach is challenging due to the nested maximizations and expectations, requiring heuristics to relax the maximization steps. By using the stochastic policy from equation \eqref{eq:sp}, we can provide a full Bayesian treatment of the dynamic programming, similar to value iteration \cite{Kaelbling96jmlr}.

\section{Results}

We show the performance of our stochastic policy for distributed BO with different acquisition functions: Expected Improvement (SP-EI), Probability of Improvement (SP-PI) and Upper Confidence Bounds (SP-UCB). We also include the parallel and distributed Thompson sampling (PDTS) \cite{hernandez2017parallel} as an alternative distributed method and the sequential expected improvement (EI) as a baseline. Note that \cite{hernandez2017parallel} already compares PDTS with parallel EI and $\epsilon$-greedy methods. In order to simplify the comparison, we use a GP as the surrogate model for all the algorithms. However, both PDTS and our stochastic policy methods allow other surrogate models such as Bayesian neural networks. In all the experiments, we assume a network of 10 nodes synchronized, that is, function evaluations are performed in batches of 10 for all distributed methods. Note that EI has an unfair advantage, as it has access to all the data for each iteration while the distributed methods only update their GP model once every 10 observations. For all the plots, we display the average of each method over 10 trials with a 95\% confidence interval. We use common random numbers to reduce the variance in the comparison and use the same initial samples among all methods.

    
    
    

\subsection{Benchmark Functions}

First, we start with a set of test problems for global optimization\footnote{From: \url{https://www.sfu.ca/~ssurjano/optimization.html}}. We have selected the functions to have a mixture of behaviours (smooth/sharp, single/multiple minima, etc.). The results on these functions are showcased in Figure \ref{benchmark}. First, we can see how the performance of SP-EI is fairly consistent among the different functions, achieving better or similar results than the alternatives. Even for functions that are difficult to optimize with BO, such as Schubert, the stochastic policy is able to perform well thanks to the extra exploration induced by the sampling process. However, this exploration does not interfere with \emph{easier} functions where exploitation is more important, such as Branin or Bohachevsky. Interestingly, SP-PI performs reasonably well. In the case of PI, the stochastic policy prevents excessive exploitation, a known behaviour of PI in the sequential BO setting \cite{shahriari2016taking}.



\begin{figure}
    \includegraphics[width=0.24\textwidth]{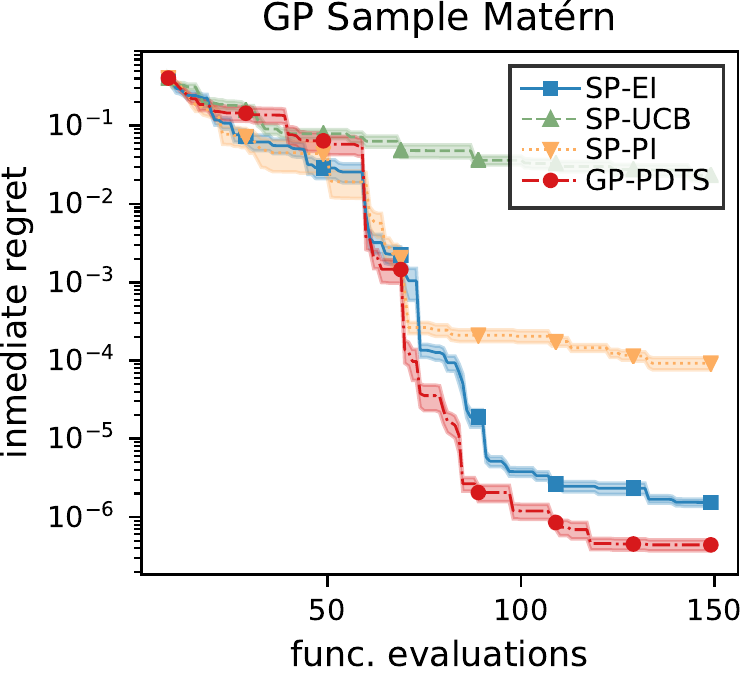}\hfill
    \includegraphics[width=0.24\textwidth]{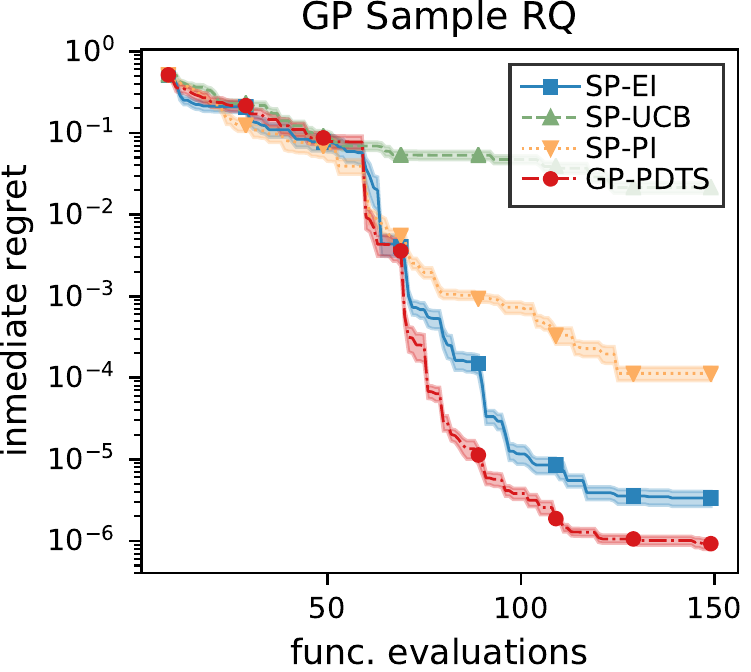}
    \caption{Optimizing random functions from a GP. A Mat\'ern kernel (left) and rational quadratic kernel (right) are used in the GP. As a Mat\'ern kernel is used in optimization, we have a \textit{within-model} (left) and \textit{out-of-model} (right) experiment. It is important to note that different kernels can generate different possible random functions, making the vertical axis not comparable between both problems.}
    \label{gpsample}
\end{figure}

We have also followed the methodology from \cite{HennigSchuler2012} and generated random functions from a known GP. We have studied two situations: a) we have studied the \textit{within-model} problem, where the GP sample uses the same kernel as the optimization algorithm (Mat\'ern with $\nu=5/2$) and, b) the \textit{out-of-model} problem where the GP sample is generated with a different kernel (a Rational Quadratic). The results in Figure \ref{gpsample} show how SP-EI is comparable to PDTS both for the \textit{within-model} and the \textit{out-of-model} problems.


\subsection{Robot Pushing}

\begin{figure}
    \includegraphics[width=0.24\textwidth]{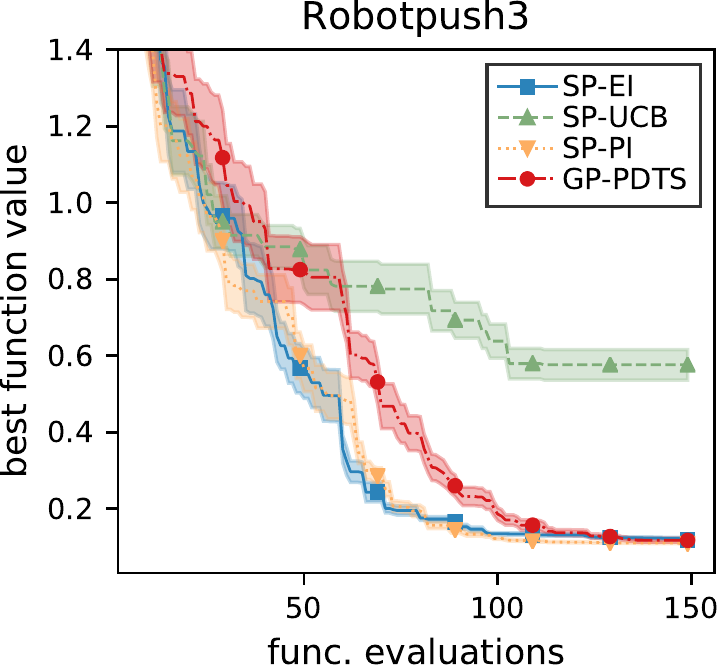}\hfill
    \includegraphics[width=0.24\textwidth]{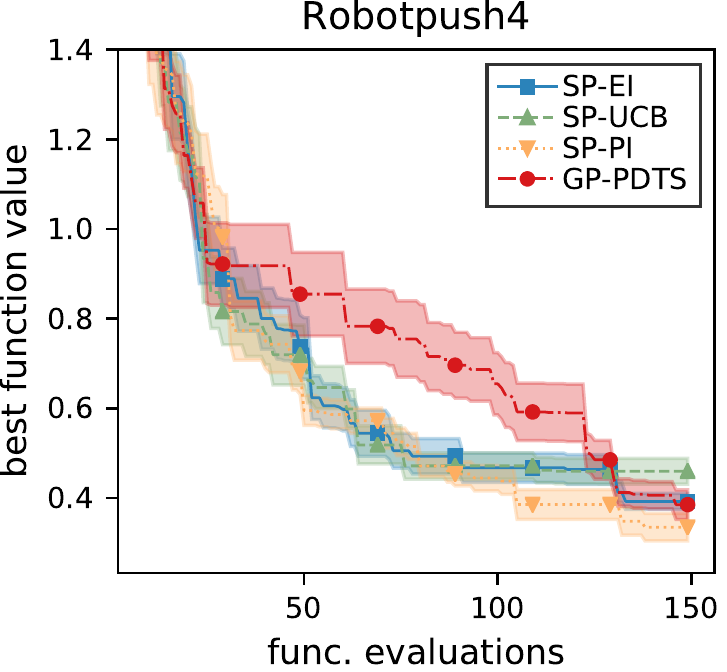}
    \caption{Robot pushing policy results, showing the 3-dimensional (left) and the 4-dimensional (right) problems.}
    \label{robotpush}
\end{figure}

In the next experiment, we use the \emph{active learning for robot pushing} setup and code from \cite{pmlr-v70-wang17e}. It consists of performing active policy search on the task of selecting a pushing action of an object towards a designated goal location. The function has a 3-dimensional input: the robot location $(r_x, r_y)$ and the pushing duration $t_r$. In a second experiment, we add the robot angle $r_\theta$ to have a 4-dimensional version. In this experiment, the repetitions are increased to 40, as each repetition is a different goal location. Figure \ref{robotpush} shows the results of both problems in which we can see how our methods SP-EI and SP-PI have faster convergence than GP-PDTS. 

\subsection{Hyperparameter Tuning of Neural Networks}

\begin{figure}
    \includegraphics[width=0.24\textwidth]{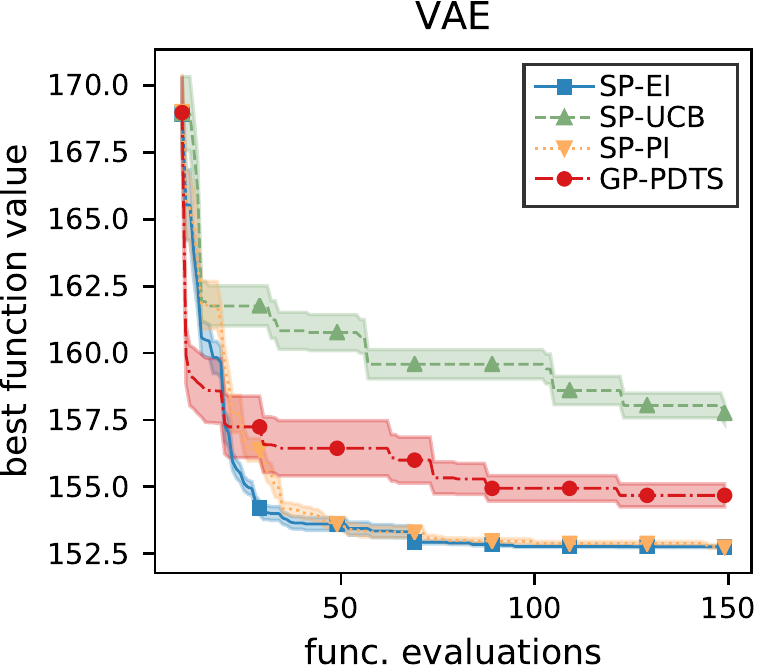}\hfill
    \includegraphics[width=0.226\textwidth]{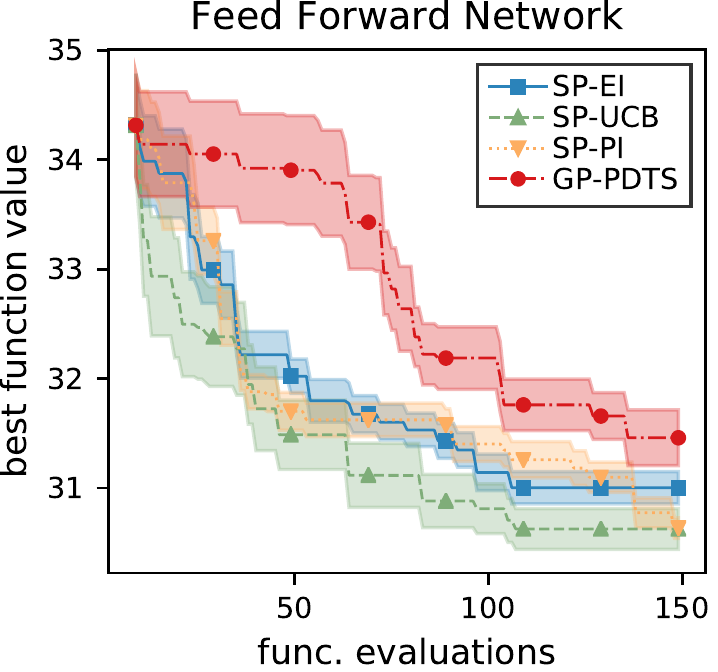}
    \caption{Hyperparameter tuning of neural network problems. A Variational Autoencoder trained with the MNIST dataset (left) and a Feedforward network trained with the Boston housing dataset (right).}
    \label{nnexamples}
\end{figure}

Finally, we use the set of problems for hyperparameter tuning of neural networks from \cite{MartinezCantin18aistats}, as these are good examples of the advantage of distributed BO for training expensive models in the cloud. We can see the results of both networks in Figure \ref{nnexamples}, where SP-EI and SP-PI consistenly outperform PDTS.

\paragraph{Variational Autoencoder (VAE) on MNIST.} A VAE is a generative method that learns a low dimensional representation of high dimensional data, such as images. We train a VAE for the MNIST dataset, and tune the following hyperparameters: number of nodes in the hidden layer, learning rate, learning rate decay and $\epsilon$ constant for the ADAM optimizer.

\paragraph{Feedforward Network on Boston Housing.} We fit a single layer feedforward network on the Boston housing dataset. The hyperparameters tuned are: number of nodes in the hidden layer, learning rate, learning rate decay and $\rho$ parameter for the exponential decay rate from RMSprop.




\section{Conclusion}

We have introduced several implications and advantages of viewing Bayesian optimization as a Markov decision process. We also have shown that this approach can be interesting for further developments of BO, both in theory and practice. As the main contribution of the paper, we have presented a new method for fully distributed BO based on stochastic policies which can be easily integrated in any setup, independent of the surrogate model or acquisition function of choice. This distributed BO allows high scalability, even by adding new resources on demand and reducing the communication between nodes. We show how, in most cases, the stochastic policy outperforms the state of the art on distributed BO (PDTS) and even to the sequential expected improvement.

\section*{Acknowledgments}
This work has been partly supported by projects DPI2015-65962-R, RTI2018-096903-B-I00 and DGA\_T45-17R.

\bibliographystyle{named}
\bibliography{ijcai19}

\end{document}